\def\year{2018}\relax
\documentclass[letterpaper]{article} 
\usepackage{aaai18}  
\usepackage{times}  
\usepackage{helvet}  
\usepackage{courier}  
\usepackage{url}  
\usepackage{graphicx}  
\usepackage{amsmath,amssymb,booktabs}
\newcommand{\E}{\mathbb{E}}
\DeclareMathOperator*{\argmax}{arg\,max}
\newcommand{\citet}[1]
{\citeauthor{#1}~\shortcite{#1}}
\newcommand{\citep}{\cite}

\begin{document}
\title{Action Branching Architectures for Deep Reinforcement Learning}
\author{Arash Tavakoli, Fabio Pardo, Petar Kormushev\\
Imperial College London\\
London SW7 2AZ, United Kingdom\\
\{a.tavakoli, f.pardo, p.kormushev\}@imperial.ac.uk
}
\maketitle

\begin{abstract}
Discrete-action algorithms have been central to numerous recent successes of deep reinforcement learning. However, applying these algorithms to high-dimensional action tasks requires tackling the combinatorial increase of the number of possible actions with the number of action dimensions. This problem is further exacerbated for continuous-action tasks that require fine control of actions via discretization. In this paper, we propose a novel neural architecture featuring a shared decision module followed by several network \textit{branches}, one for each action dimension. This approach achieves a linear increase of the number of network outputs with the number of degrees of freedom by allowing a level of independence for each individual action dimension. To illustrate the approach, we present a novel agent, called Branching Dueling Q-Network (BDQ), as a branching variant of the Dueling Double Deep Q-Network (Dueling DDQN). We evaluate the performance of our agent on a set of challenging continuous control tasks. The empirical results show that the proposed agent scales gracefully to environments with increasing action dimensionality and indicate the significance of the shared decision module in coordination of the distributed action branches. Furthermore, we show that the proposed agent performs competitively against a state-of-the-art continuous control algorithm, Deep Deterministic Policy Gradient (DDPG).   
\end{abstract}

\section{Introduction}
\label{sec:intro}

Combining the recent advances in deep learning techniques \citep{Lecun:2015deep,Schmidhuber:2015NN,Goodfellow:2016DLbook} with reinforcement learning algorithms \citep{Bertsekas:1996neuro,Sutton:1998RLbook,Szepesvari:2010algorithms} has proven to be effective in many domains. Notable examples include the Deep Q-Network (DQN) \citep{Mnih:2013,Mnih:2015natureDQN} and AlphaGo \citep{Silver:2016AlphaGo,Silver:2017AlphaGoZero}. The main advantage of using neural networks as function approximators in reinforcement learning is their ability to deal with high-dimensional input data by modeling complex hierarchical or compositional data abstractions and features.

Despite these successes, which have enabled the use of reinforcement learning algorithms in domains with unprocessed, high-dimensional sensory input, the application of these methods to high-dimensional, discrete action spaces remains to suffer from the same issues as in tabular reinforcement learning---that is, the number of actions that need to be explicitly represented grows exponentially with increasing action dimensionality. Formally, for an environment with an $N$-dimensional action space and $n_d$ discrete sub-actions for each dimension $d$, using the existing discrete-action algorithms, a total of $\prod_{d=1}^{N}{n_d}$ possible actions need to be considered. This can rapidly render the application of discrete-action reinforcement learning algorithms intractable to domains with multidimensional action spaces, as such large action spaces are difficult to explore efficiently \citep{Lillicrap:2016ddpg}. This limitation is a significant one as there are numerous efficient discrete-action algorithms whose applications are currently restricted to domains with relatively small discrete action spaces. For instance, Q-learning \citep{Watkins:1992Q} is a powerful discrete-action algorithm, with many extensions \citep{Hessel:2017Rainbow}, which due to its off-policy nature can, in principle, achieve better sample efficiency than policy gradient methods by reusing transitions from a replay memory of past experience transitions or demonstrations \citep{Gu:2016naf,Gu:2017qprop}.

\begin{figure}[t!]
  \centering
  \includegraphics[width=1.0\linewidth]{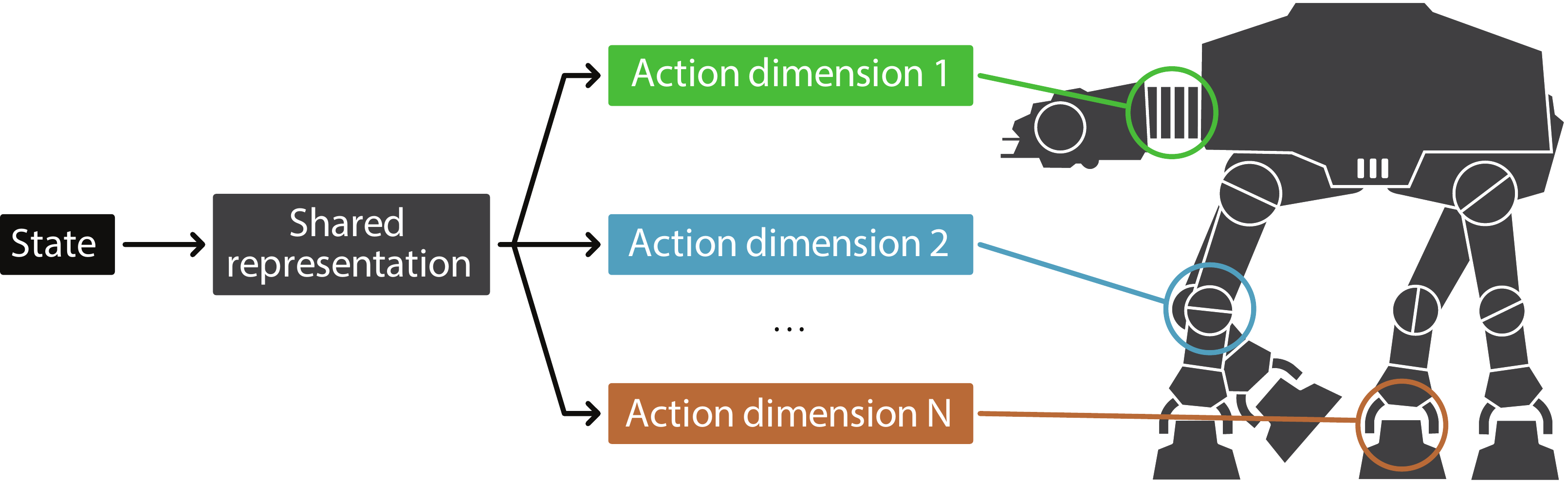}
  \caption{A conceptual illustration of the proposed action branching network architecture. The shared network module computes a latent representation of the input state that is then passed forward to the several action branches. Each action branch is responsible for controlling an individual degree of freedom and the concatenation of the selected sub-actions results in a joint-action tuple.}
\label{fig:action_branching_architectures}
\end{figure}

Given the potential of discrete-action reinforcement learning algorithms and their current limited application, in this paper we introduce a novel neural architecture that enables the use of discrete-action algorithms in deep reinforcement learning for domains with high-dimensional discrete or continuous action spaces. The core notion of the proposed architecture is to distribute the representation of the action controllers across individual network \textit{branches}, meanwhile, maintaining a \textit{shared decision module} among them to encode a latent representation of the input and help with the coordination of the branches (see Figure~\ref{fig:action_branching_architectures}). The proposed decomposition of the actions enables the linear growth of the total number of network outputs with increasing action dimensionality as opposed to the combinatorial growth in current discrete-action algorithms. This simple idea can potentially enable a spectrum of fundamental discrete-action reinforcement learning algorithms to be effectively applied to domains with high-dimensional discrete or continuous action spaces using neural network function approximators. 

To showcase this capability, we introduce a novel agent, called Branching Dueling Q-Network (BDQ), which is a branching variant of the Dueling Double DQN (Dueling DDQN) \citep{Wang:2016}. We evaluate BDQ on a variety of complex control problems via fine-grained discretization of the continuous action space. Our empirical study shows that BDQ can scale robustly to environments with high-dimensional action spaces to solve the benchmark domains and even outperform the Deep Deterministic Policy Gradient (DDPG) algorithm \citep{Lillicrap:2016ddpg} in the most challenging task with a corresponding discretized combinatorial action space of approximately $6.5 \times 10^{25}$ action tuples. To solve problems in environments with discrete action spaces of this magnitude is a feat that was previously thought intractable for discrete-action algorithms \citep{Lillicrap:2016ddpg,Schulman:2017ppo}. In order to demonstrate the vital role of the shared decision module in our architecture, we compare BDQ against a completely independent variant which we refer to as Independent Dueling Q-Network (IDQ)---an agent consisting of multiple independent networks, one for each action dimension, and without any shared parameters among the networks. The results show that the performance of IDQ quickly deteriorates with increasing action dimensionality. This implies the inability of the agent to coordinate the independent action decisions across its several networks. This could be due to any of the several well-known issues for independent fully-cooperative learning agents: Pareto-selection, non-stationarity, stochasticity, alter-exploration, and shadowed equilibria \citep{Matignon:2012independent}.

Partial distribution of control, or \textit{action branching} as we call it, is also found in nature. Octopuses, for instance, have complex neural systems where each arm is able to function with a degree of autonomy and even respond to stimuli after being detached from the central control. In fact, more than half of the neurons in an octopus are spread throughout its body, especially within the arms \citep{Godfrey:2016octopus}. Since the octopuses' arms have virtually unlimited degrees of freedom, they are highly difficult to control in comparison to jointed limbs. 
This calls for the partial delegation of control to the arms in order to work out the details of their motions themselves. Interestingly, not only do the arms have a degree of autonomy, they have also been observed to engage in independent exploration \citep{Godfrey:2016octopus}.

\section{Related Work}
\label{sec:relatedwork}

To enable the application of reinforcement learning algorithms to large-scale, discrete-action problems, \citet{Dulac:2015} propose the Wolpertinger policy architecture based on a combination of DDPG and an approximate nearest-neighbor method. This approach leverages prior information about the discrete actions in order to embed them in a continuous space upon which it can generalize, meanwhile, achieving logarithmic-time lookup complexity relative to the number of actions. Due to the underlying algorithm being essentially a continuous-action algorithm, this approach may be unsuitable for domains with naturally discrete action spaces where no assumption should be imposed on having associated continuous space correlations. Also, this approach does not enable the application of discrete-action algorithms to domains with high-dimensional action spaces as it relies on a continuous-action algorithm.

Concurrent to our work, \citet{Metz:2017} have developed an approach that can deal with problem domains with high-dimensional discrete action spaces using Q-learning. They use an autoregressive network architecture to sequentially predict the action value for each action dimension. This requires manual ordering of the action dimensions which imposes \textit{a priori} assumptions on the structure of the task. Additionally, due to the sequential structure of the network, as the number of action dimensions increases, so does the noise in the Q-value estimations. Therefore, with increasing number of action dimensions, the Q-value estimates on the latter layers may become too noisy to be useful. Due to the parallel representation of the action values or policies, our proposed approach is not prone to cumulative estimation noise with increasing number of action dimensions and does not impose manual action factorization. Furthermore, our proposed approach is much simpler to implement as it does not require advanced neural network architectures, such as recurrent neural networks.

A potential approach towards achieving scalability with increasing number of action dimensions is to extend deep reinforcement learning algorithms to fully-cooperative multiagent settings in which each agent---responsible for controlling an individual degree of freedom---observes the global state, selects an individual action, and receives a team reward common to all agents. \citet{Tampuu:2017independentDQN} combine DQN with independent Q-learning, in which each agent independently and simultaneously learns its own action-value function. Even though this approach has been successfully applied in practice to domains with two agents, in principle, it can lead to convergence problems \citep{Matignon:2012independent}. In this paper, we empirically investigate this scenario and show that by maintaining a shared set of parameters among the action branches, our proposed approach is able to scale to high-dimensional action spaces.  

\begin{figure*}[t!]
  \centering
  \includegraphics[width=.75\linewidth]{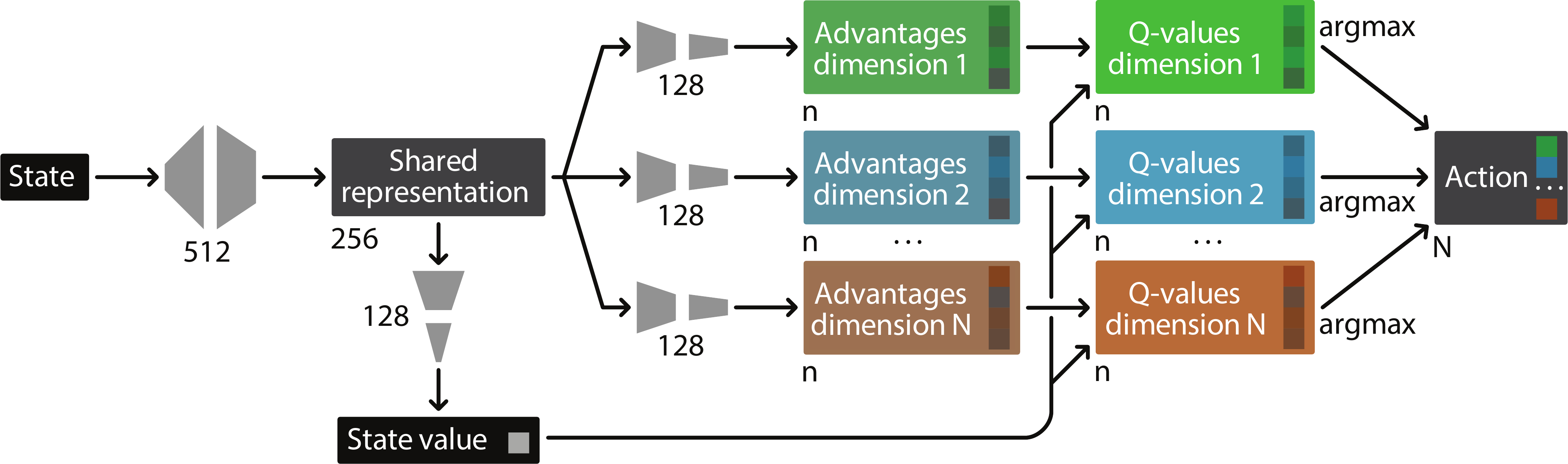}
  \caption{A visualization of the specific action branching network implemented for the proposed BDQ agent. When a state is provided at the input, the shared decision module computes a latent representation that is then used for evaluation of the state value and the factorized (state-dependent) action advantages on the subsequent independent branches. The state value and the factorized advantages are then combined, via a special aggregation layer, to output the Q-values for each action dimension. These factorized Q-values are then queried for the generation of a joint-action tuple. The weights of the fully connected neural layers are denoted by the gray trapezoids and the size of each layer (i.e. number of units) is indicated.}
\label{fig:bdq_network}
\end{figure*}

\section{Action Branching Architecture}
\label{sec:method}

The key insight behind the proposed action branching architecture is that for solving problems in multidimensional action spaces, it is possible to optimize for each action dimension with a degree of independence. If executed appropriately, this altered perspective has the potential to trigger a dramatic reduction in the number of required network outputs. However, it is well-known that the na{\"i}ve distribution of the value function or the policy representation across several independent function approximators is subject to numerous challenges which can lead to convergence problems \citep{Matignon:2012independent}.
To address this, the proposed neural architecture distributes the representation of the value function or the policy across several network branches while keeping a shared decision module among them to encode a latent representation of the common input state (see Figure~\ref{fig:action_branching_architectures}). We hypothesize that this shared network module, paired with an appropriate training procedure, can play a significant role in coordinating the sub-actions that are based on the semi-independent branches and, therefore, achieve training stability and convergence to good policies. We believe this is due to the rich features in the shared network module that is trained via the backpropagation of the gradients originating from all the branches.

To verify this capability, we present a novel agent that is based on the incorporation of the proposed action branching architecture into a popular discrete-action reinforcement learning agent, the Dueling Double Deep Q-Network (Dueling DDQN). The proposed agent, which we call Branching Dueling Q-Network (BDQ), is only an example of how we envision our action branching architecture can be combined with a discrete-action algorithm in order to enable its direct application to problem domains with high-dimensional, discrete or continuous action spaces. We select deep Q-learning (also known as DQN) as the algorithmic basis for our proof-of-concept agent as it is a simple, yet powerful, off-policy algorithm with an excellent track record and numerous extensions \citep{Hessel:2017Rainbow}. 

While our experiments focus on a specific algorithm (i.e. deep Q-learning), we believe that the empirical verification of the aforementioned hypothesis, suggests the potential of the proposed approach in enabling the direct application of a spectrum of existing discrete-action algorithms to environments with high-dimensional action spaces.

\section{Branching Dueling Q-Network}
\label{sec:bdq}

In this section, we begin by providing a brief overview of a select set of available extensions for DQN that we incorporate into the proposed BDQ agent. We then describe the details of the proposed agent, including the specific methods that were used to adapt the DQN algorithm and its selected extensions into our proposed action branching architecture. Figure~\ref{fig:bdq_network} demonstrates a pictorial view of the BDQ network.

\subsection{Background}
\label{subsec:background}

The following is an outline of three existing key innovations, designed to improve upon the sample efficiency and policy evaluation quality of the DQN algorithm. 

\paragraph{Double Q-learning} Both tabular Q-learning and DQN have been shown to suffer from the overestimation of the action values \citep{Hasselt:2010dq,Hasselt:2016ddq}. This overoptimism stems from the fact that the same values are accessed in order to both select and evaluate actions. In the standard DQN algorithm \citep{Mnih:2013,Mnih:2015natureDQN}, a previous version of the current Q-network, called the target network, is used to select the next greedy action involved in the Q-learning updates. To address the overoptimism in the Q-value estimations, \citet{Hasselt:2016ddq} propose the Double DQN (DDQN) algorithm that uses the current Q-network to select the next greedy action, but evaluates it using the target network.

\paragraph{Prioritized Experience Replay} The experience replay enables online, off-policy reinforcement learning agents to reuse past experiences or demonstrations. In the standard DQN algorithm, the experience transitions were sampled uniformly from a replay buffer. To enable more efficient learning from the experience transitions, \citet{Schaul:2016prior} propose a framework for prioritizing experience in order to replay important experience transitions, which have a high expected learning progress, more frequently.

\paragraph{Dueling Network Architecture} The dueling network architecture \citep{Wang:2016} explicitly separates the representation of the state value and the (state-dependent) action advantages into two separate branches while sharing a common feature-learning module among them. The two branches are combined, via a special aggregating layer, to produce an estimate of the action-value function. By training this network with no additional considerations than those used for the DQN algorithm, the dueling network automatically produces separate estimates of the state value and advantage functions. \citet{Wang:2016} introduce multiple aggregation methods for combining the state value and advantages. They demonstrate that subtracting the mean of the advantages from each individual advantage and then summing them with the state value results in improved learning stability when compared to the na{\"i}ve summation of the state value and advantages. The dueling network architecture has been shown to lead to better policy evaluation in the presence of many similar-valued (or redundant) actions, and thus achieves faster generalization over large action spaces.

\subsection{Methods}
\label{subsec:bdq_methods}

Here we introduce various methods for adapting the DQN algorithm, as well as its notable extensions that were explained earlier, into the action branching architecture. For brevity, we mainly focus on the methods that result in our best performing DQN-based agent, BDQ.

\paragraph{Common State-Value Estimator} As demonstrated in the action branching network of Figure~\ref{fig:bdq_network}, BDQ uses a common state-value estimator for all action branches. This approach, which can be thought of as an adaptation of the dueling network into the action branching architecture, generally yields a better performance. The use of the dueling architecture with action branching is particularly an interesting augmentation for learning in large action spaces. This is due to the fact that the dueling architecture can more rapidly identify action redundancies and generalize more efficiently by learning a general value that is shared across many similar actions. In order to adapt the dueling architecture into our action branching network, we distribute the representation of the (state-dependent) action advantages on the several action branches, meanwhile, adding a single additional branch for estimating the state-value function. Similar to the dueling architecture, the advantages and the state value are combined, via a special aggregating layer, to produce estimates of the distributed action values. We experimented with several aggregation methods and our best performing method is to locally subtract each branch's mean advantage from its sub-action advantages, prior to their summation with the state value. Formally, for an action dimension $d \in \{1,...,N\}$ with $|\mathcal{A}_d| = n$ discrete sub-actions, the individual branch's Q-value at state $s \in \mathcal{S}$ and sub-action $a_d \in \mathcal{A}_d$ is expressed in terms of the common state value $V(s)$ and the corresponding (state-dependent) sub-action advantage $A_d(s,a_d)$ by:
\begin{equation}
    Q_d(s,a_d) = V(s) + \big( A_d(s,a_d) - \frac{1}{n} \sum_{a'_d \in \mathcal{A}_d} A_d(s,a'_d) \big).
\label{equ:agg_reduce_mean}
\end{equation}

We realize that this aggregation method does not resolve the lack of identifiability for which the maximum and the average reduction methods were originally proposed \citep{Wang:2016}. However, based on our experimentation, this aggregation method yields a better performance than both the na{\"i}ve alternative, 
\begin{equation}
    Q_d(s,a_d) = V(s) + A_d(s,a_d), 
\label{equ:agg_naive}
\end{equation}
and the local maximum reduction method, which replaces the averaging operator in Equation~\ref{equ:agg_reduce_mean} with a maximum operator:
\begin{equation}
    Q_d(s,a_d) = V(s) + \big( A_d(s,a_d) - \max_{a'_d \in \mathcal{A}_d} A_d(s,a'_d) \big).
\label{equ:agg_reduce_max}
\end{equation}

\paragraph{Temporal-Difference~Target} We tried several different methods for generating the temporal-difference (TD) targets for the DQN updates. A simple approach is to calculate a TD target, similar to that in DDQN, for each individual action dimension separately:
\begin{equation}
    y_d = r + \gamma Q^-_d(s',\argmax_{a'_d \in \mathcal{A}_d} Q_d(s',a'_d)),
\label{equ:TD_target_DDQN}
\end{equation}
with $Q^-_d$ denoting the branch $d$ of the target network $Q^-$. 

Alternatively, the maximum DDQN-based TD target over the action branches may be set as a single global learning target for all action dimensions:
\begin{equation}
    y = r + \gamma \max_d Q^-_d(s',\argmax_{a'_d \in \mathcal{A}_d} Q_d(s',a'_d)).
\label{equ:TD_target_max}
\end{equation}

The best performing method, also used for BDQ, replaces the maximum operator in Equation~\ref{equ:TD_target_max} with a mean operator:  
\begin{equation}
    y = r + \gamma \frac{1}{N} \sum_d Q^-_d(s',\argmax_{a'_d \in \mathcal{A}_d} Q_d(s',a'_d)).
\label{equ:TD_target_mean}
\end{equation}

\paragraph{Loss Function} There exist numerous ways by which the distributed TD errors across the branches can be aggregated to specify a loss. A simple approach is to define the loss to be the expected value of a function of the averaged TD errors across the branches. However, due to the signs of such errors, their summation is subject to canceling out which, in effect, generally reduces the magnitude of the loss. To overcome this, the loss can be specified as the expected value of a function of the averaged absolute TD errors across the branches. In practice, we found that defining the loss to be the expected value of the mean squared TD error across the branches mildly enhances the performance: 
\begin{equation}
    L = \E_{(s,a,r,s') \sim \mathcal{D}} \left[ \frac{1}{N} \sum_d \big(y_d - Q_d(s,a_d) \big)^{2} \right],
\label{equ:loss}
\end{equation}
where $\mathcal{D}$ denotes a (prioritized) experience replay buffer and $a$ denotes the joint-action tuple $(a_1,a_2,...,a_N)$. 

\paragraph{Gradient Rescaling} During the backward pass, since all branches backpropagate gradients through the shared network module, we rescale the combined gradient prior to entering the deepest layer in the shared network module by $1/{(N + 1)}$.   

\begin{figure*}[ht] 
  \centering
  \includegraphics[height=32.75mm]{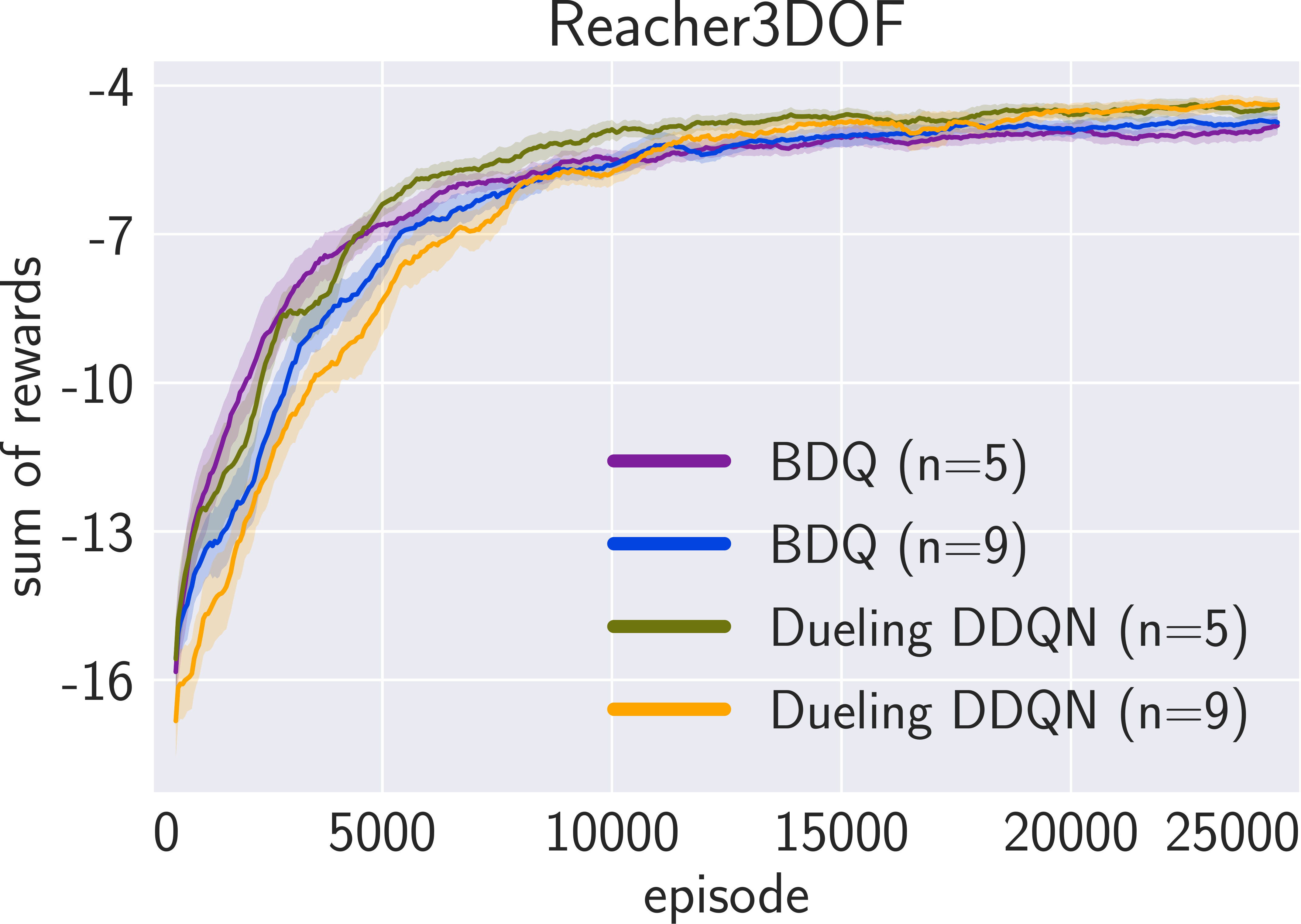} \hspace{3.8mm}
  \includegraphics[height=32.75mm]{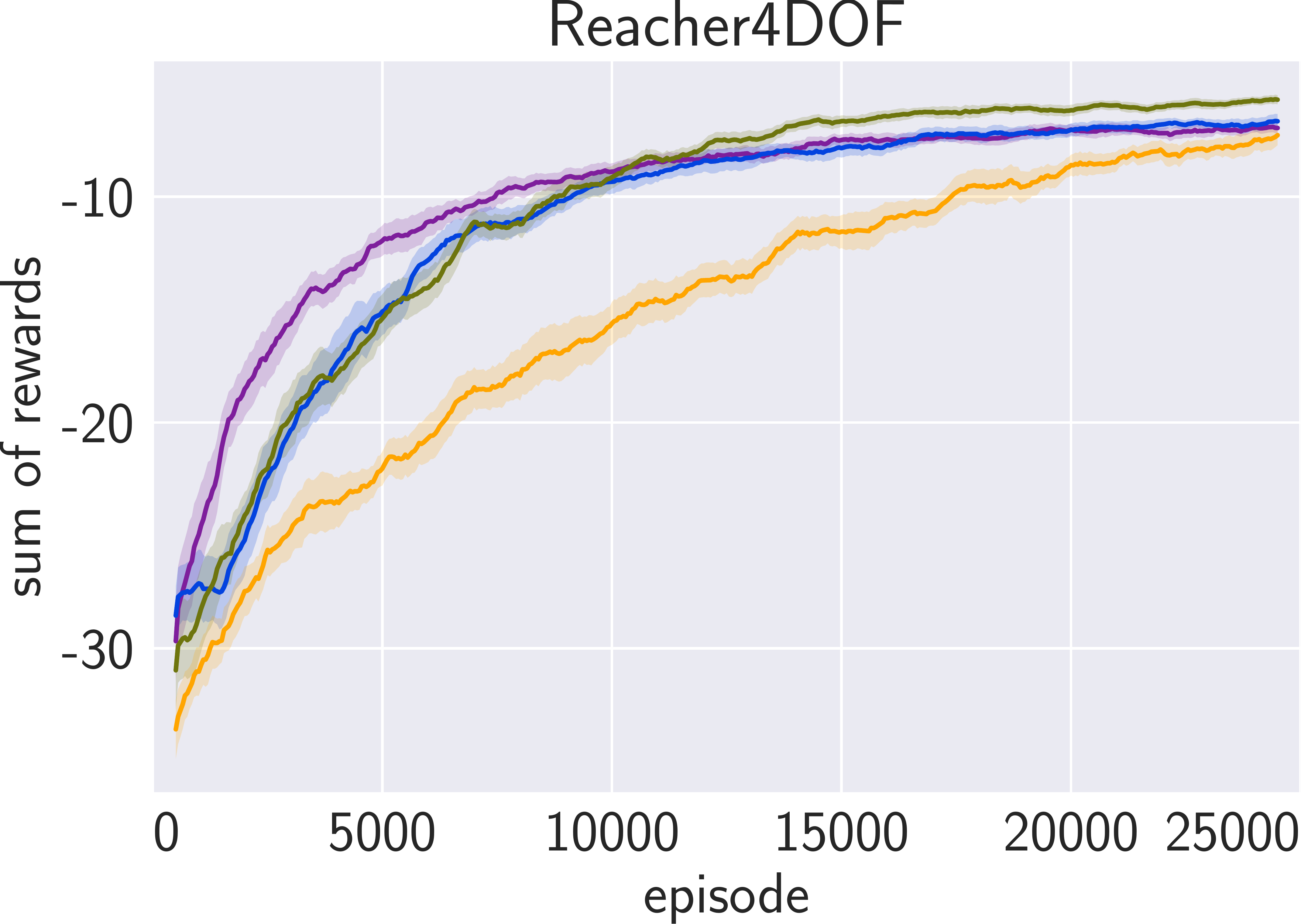} \hspace{3.8mm}
  \includegraphics[height=32.75mm]{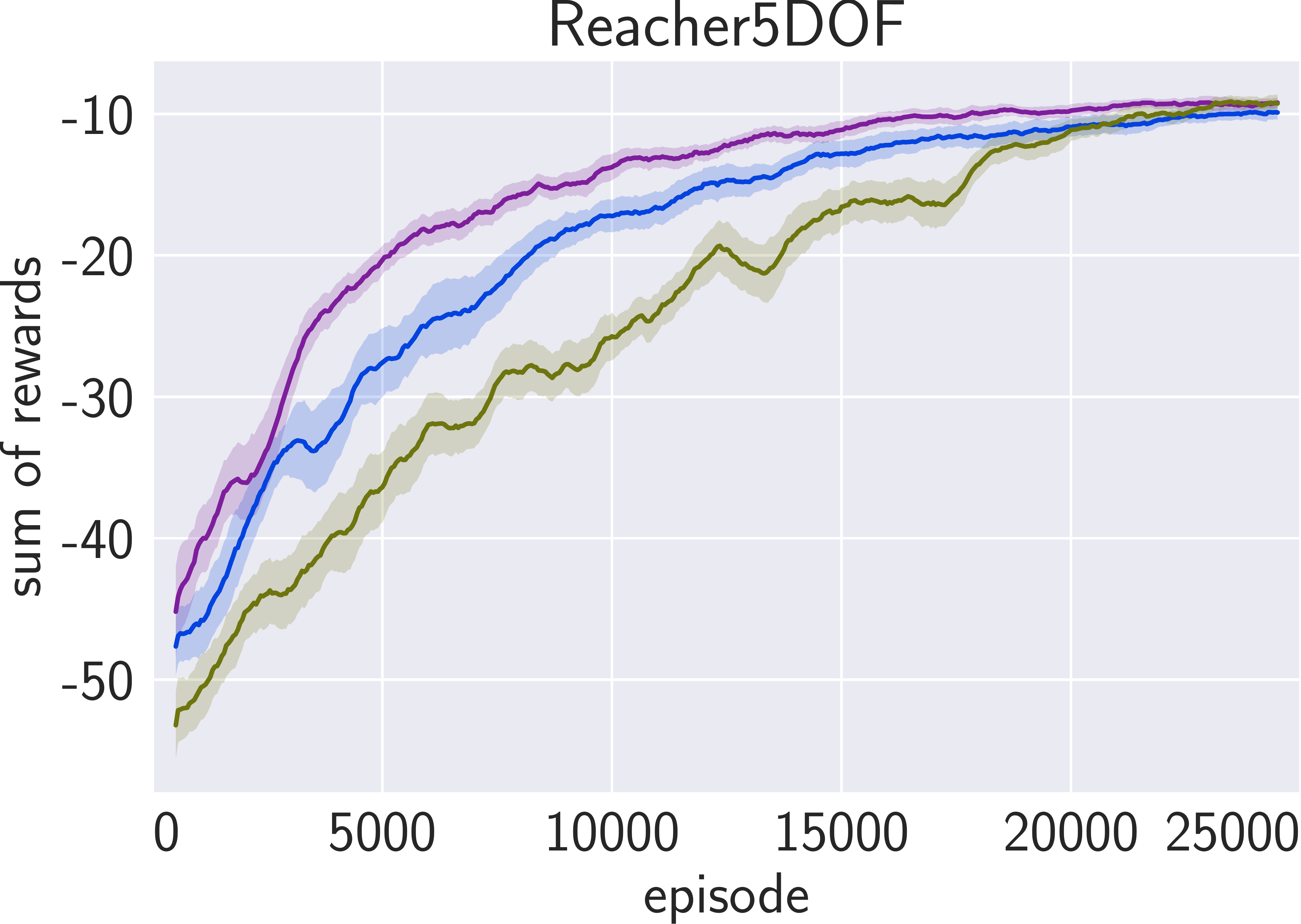}
  \caption{Performance in sum of rewards during evaluation on the $y$-axis and training episodes on the $x$-axis. The solid lines represent smoothed (window size of $20$ episodes) averages over $3$ runs with random initialization seeds, while shaded areas show the standard deviations. Evaluations were conducted every $50$ episodes of training for $30$ episodes with a greedy policy.}
\label{fig:results_custom_reachers}
\end{figure*}

\paragraph{Error for Experience Prioritization} Adapting the prioritized replay into the action branching architecture requires an appropriate method for aggregating the distributed TD errors (of a single transition) into a unified one. This error is then used by the replay memory to calculate the transition's priority. In order to preserve the magnitudes of the errors, for BDQ, we specify the unified prioritization error to be the sum across a transition's absolute, distributed TD errors:
\begin{equation}
    e_D(s,a,r,s') = \sum_d \left|y_d - Q_d(s,a_d) \right|,
\label{equ:PER_error}
\end{equation}
where $e_D(s,a,r,s')$ denotes the error used for prioritization of the experience transition tuple $(s,a,r,s')$.

\section{Experiments}
\label{sec:experiments}

We evaluate the performance of the proposed BDQ agent on several challenging continuous control environments of varying action dimensionality and complexity. These environments are simulated using the MuJoCo physics engine \citep{Todorov:2012mujoco}. We first study the performance of BDQ against its standard non-branching variant, Dueling DDQN, on a set of custom reaching tasks with increasing degrees of freedom and under two different granularity discretizations. We then compare the performance of BDQ against a state-of-the-art continuous control algorithm, Deep Deterministic Policy Gradient (DDPG), on a set of standard continuous control manipulation and locomotion benchmark domains from the OpenAI's MuJoCo Gym collection \citep{Brockman:2016gym,Duan:2016benchmarking}. We also compare BDQ against a fully independent alternative, Independent Dueling Q-Network (IDQ), in order to verify our hypothesis regarding the significance of the shared network module in coordinating the distributed policies. To make the continuous-action domains compatible with the discrete-action algorithms in our study (i.e. BDQ, Dueling DDQN, and IDQ), in both sets of experiments, we discretize each action dimension $d \in \{1,...,N\}$, in the underlying continuous action space, into $n$ equally spaced values, yielding a discrete combinatorial action space of $n^N$ possible actions.

\subsection{Custom Reaching Environments}
\label{sec:experiments_arms}

\begin{figure}[t!]
  \centering
  \includegraphics[width=.27\linewidth]{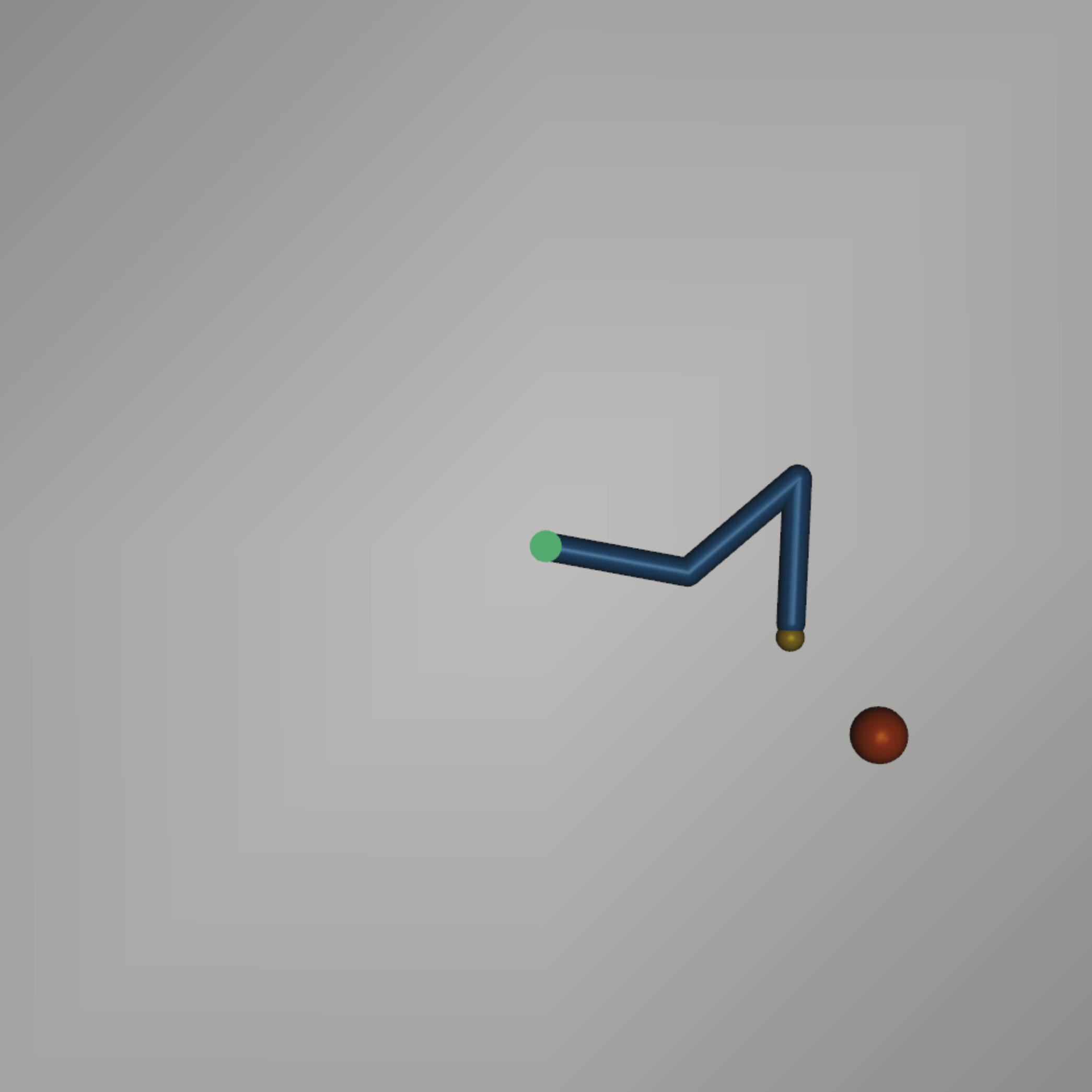}
  \includegraphics[width=.27\linewidth]{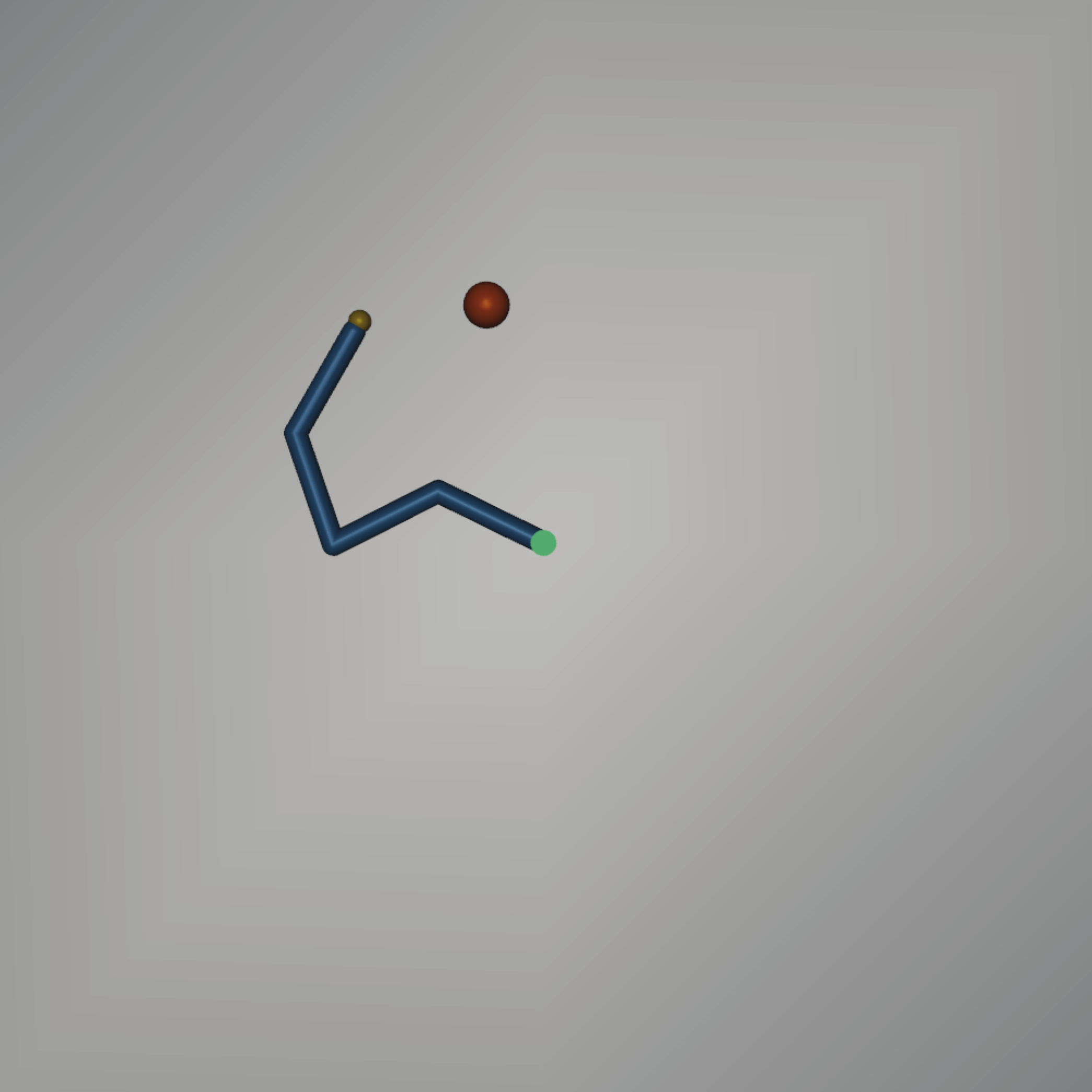}
  \includegraphics[width=.27\linewidth]{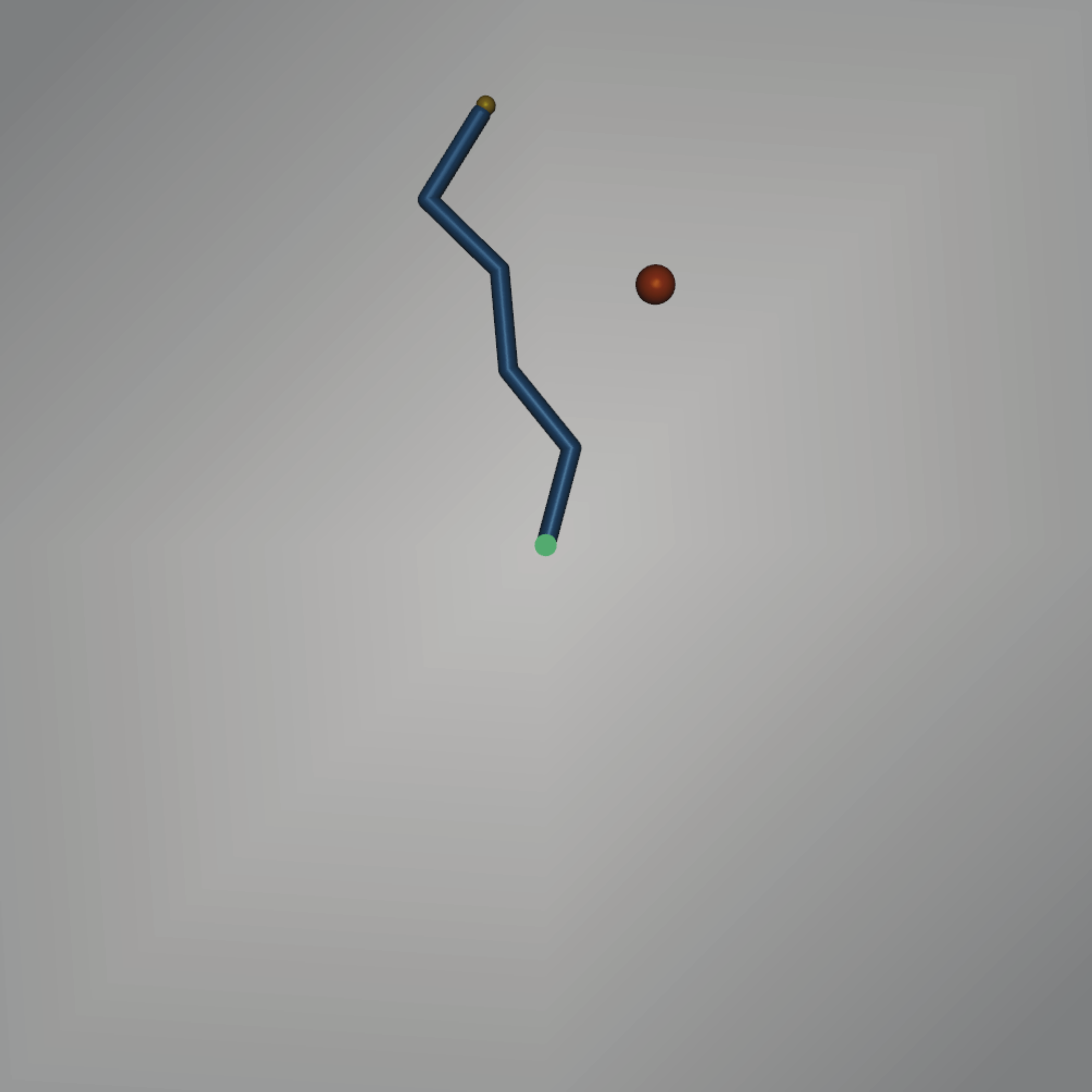}
  \caption{Illustrations of the custom physical reaching tasks. From left: Reacher3DOF, Reacher4DOF, and Reacher5DOF domains with $3$, $4$, and $5$ degrees of freedom, respectively.}
\label{fig:illustrations_custom_reachers}
\end{figure}

We begin by comparing the performance of BDQ against its standard non-branching variant, the Dueling DDQN agent, on a set of physical manipulation tasks with increasing action dimensionality (see Figure~\ref{fig:illustrations_custom_reachers}). These tasks are custom variants of the standard Reacher-v1 task (from the OpenAI's MuJoCo Gym collection) that feature more actuated joints (i.e. $N = \{3,4,5\}$) with constraints on their ranges of motion to prevent collision between segments. Unlike the original Reacher-v1 domain, reaching the target position immediately terminates an episode without the need to decelerate and maintain position at the target. This was done to simplify these complex control tasks (as a result of more frequently experienced episodic successes) in order to allow faster experimentation. We consider two discretization resolutions resulting in $n=5$ and $n=9$ sub-actions per joint. This is done in order to examine the impact of finer granularity, or equivalently more discrete sub-actions per action dimension, with increasing degrees of freedom. The general idea is to empirically study the effectiveness of action branching in the face of increasing action-space dimensionality as compared to the standard non-branching variant. Therefore, the tasks are designed to have sufficiently small action spaces for a standard non-branching algorithm to still be tractable for the purpose of our evaluations.

\begin{figure*}[ht] 
  \centering
  \includegraphics[height=32.75mm]{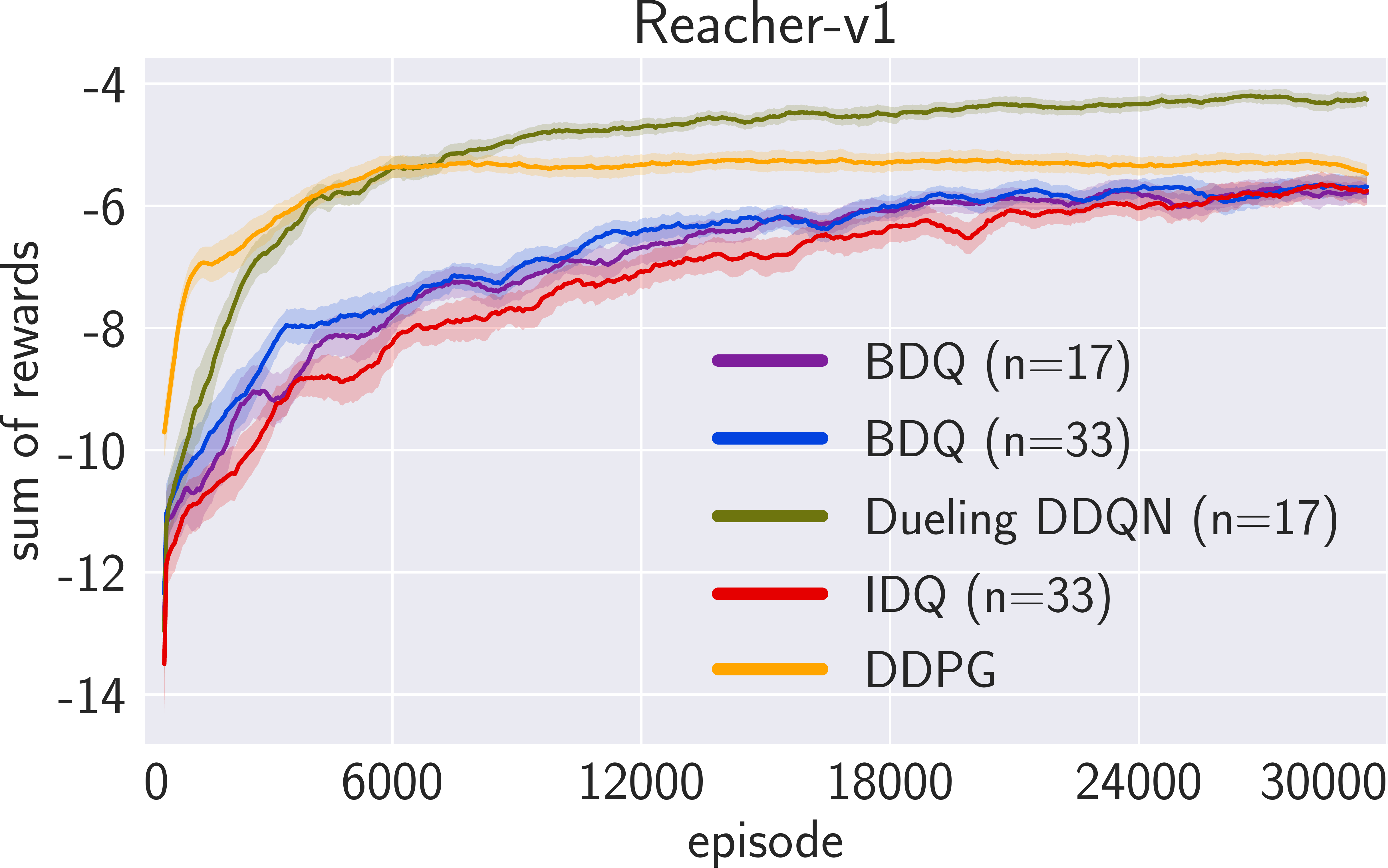} \hspace{3.8mm}
  \includegraphics[height=32.75mm]{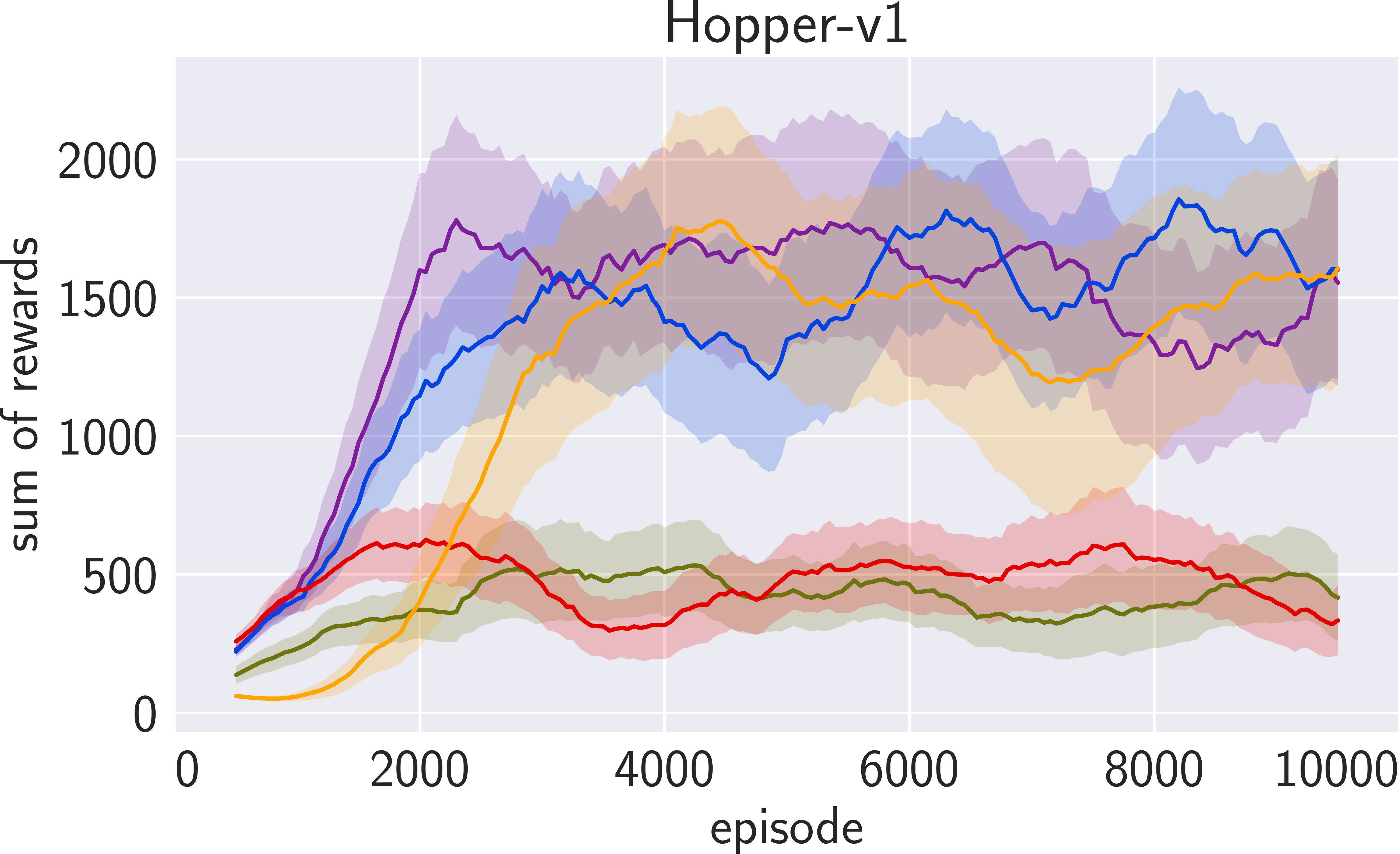}\\ \vspace{.2mm}
  \includegraphics[height=32.75mm]{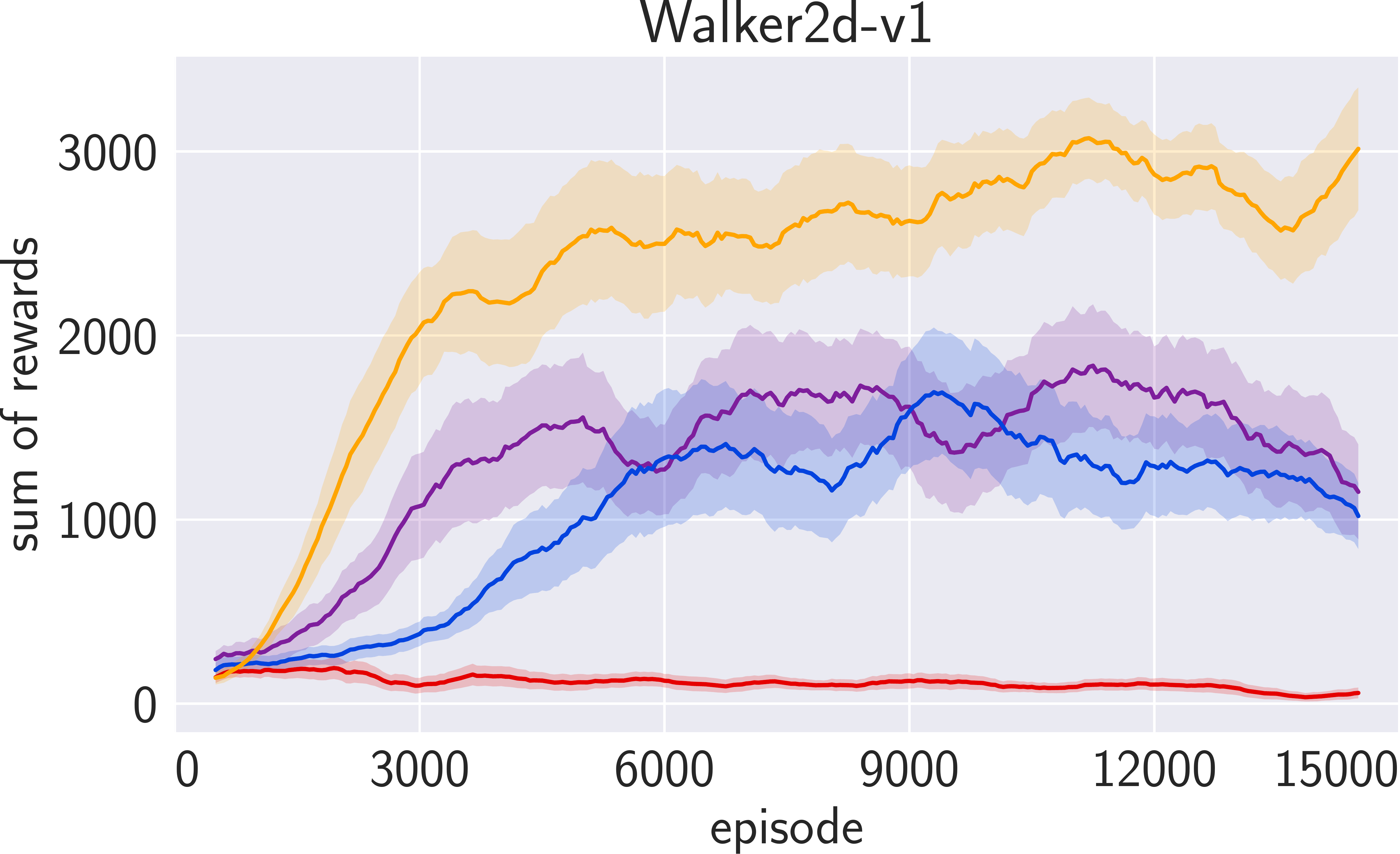} \hspace{3.8mm}
  \includegraphics[height=32.75mm]{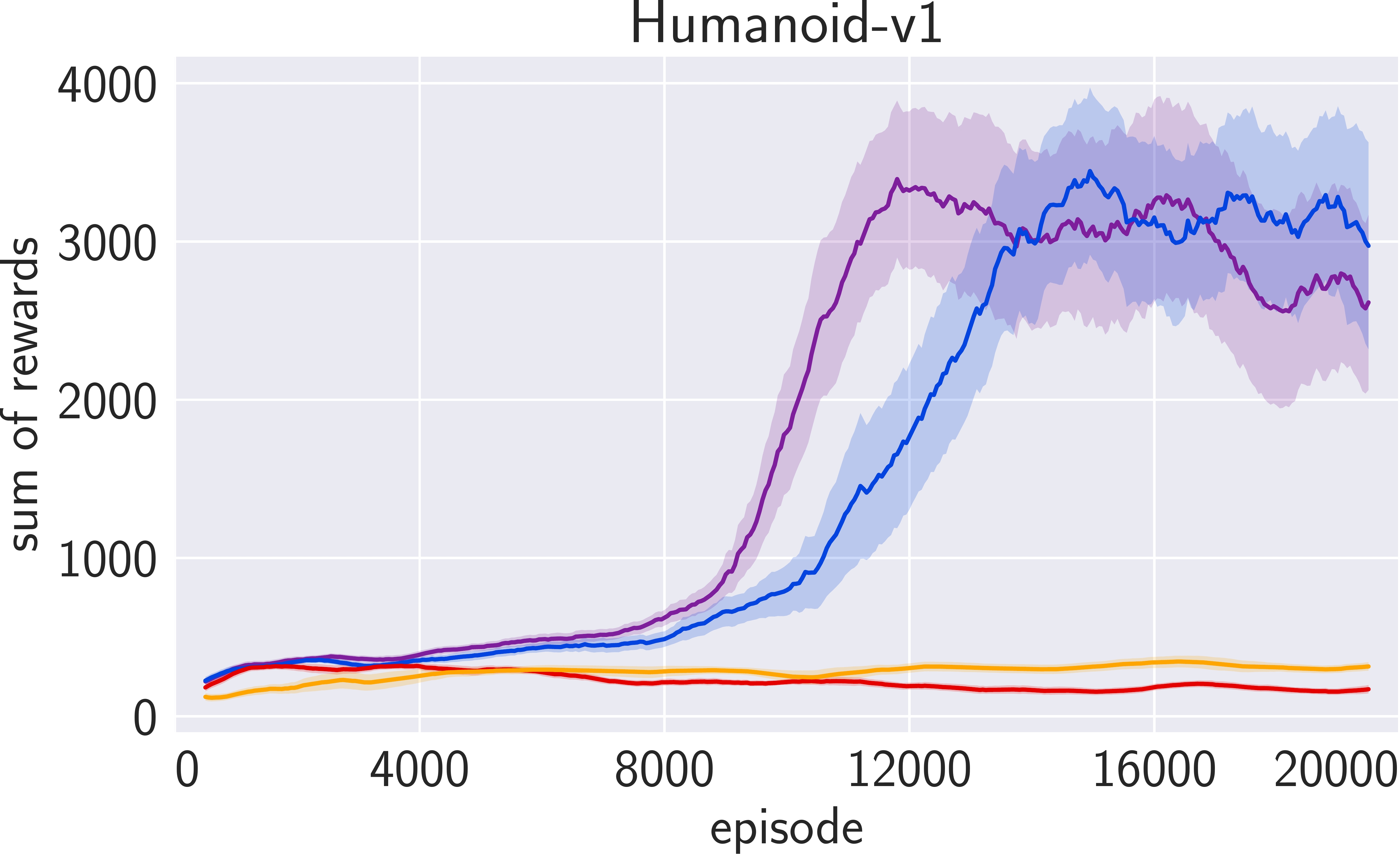}
  \caption{Learning curves for the OpenAI's MuJoCo Gym manipulation and locomotion benchmark domains. The solid lines represent smoothed (window size of $20$ episodes) averages over $6$ runs with random initialization seeds, while shaded areas show the standard deviations. Evaluations were conducted every $50$ episodes of training for $30$ episodes with a greedy policy.}
\label{fig:results_OpenAI_benchmarks}
\end{figure*}

The performances are summarized in Figure~\ref{fig:results_custom_reachers}. The results show that in the low-dimensional reaching task with $N=3$, all agents learn at about the same rate, with slightly steeper learning curves towards the end for Dueling DDQN. In the task with $N=4$, we see that the Dueling DDQN agent with $n=5$ starts off less efficiently (i.e. slower learning curve) than its corresponding BDQ agent, but eventually converges and outperforms both BDQ agents in their final performance. However, in the same task, the Dueling DDQN agent with $n=9$ shows a significantly less efficient learning performance against its BDQ counterpart. In the high-dimensional reaching task with $N=5$, we see that the Dueling DDQN agent with $n=5$ performs rather poorly in terms of its sample efficiency. For this task, we were unable to run the Dueling DDQN agent with $n=9$ since running it was computationally expensive---due to the large number of actions that need to be explicitly represented by its network (i.e. $9^5 \approx 6 \times 10^4$) and consequently the extremely large number of network parameters that need to be trained at every iteration. In contrast, in the same task, we see that BDQ performs well and converges to good policies with robustness against the discretization granularity.

\subsection{Standard Benchmark Environments}
\label{sec:experiments_openai}

\begin{figure}[t]
  \centering
  \includegraphics[width=.2428\linewidth]{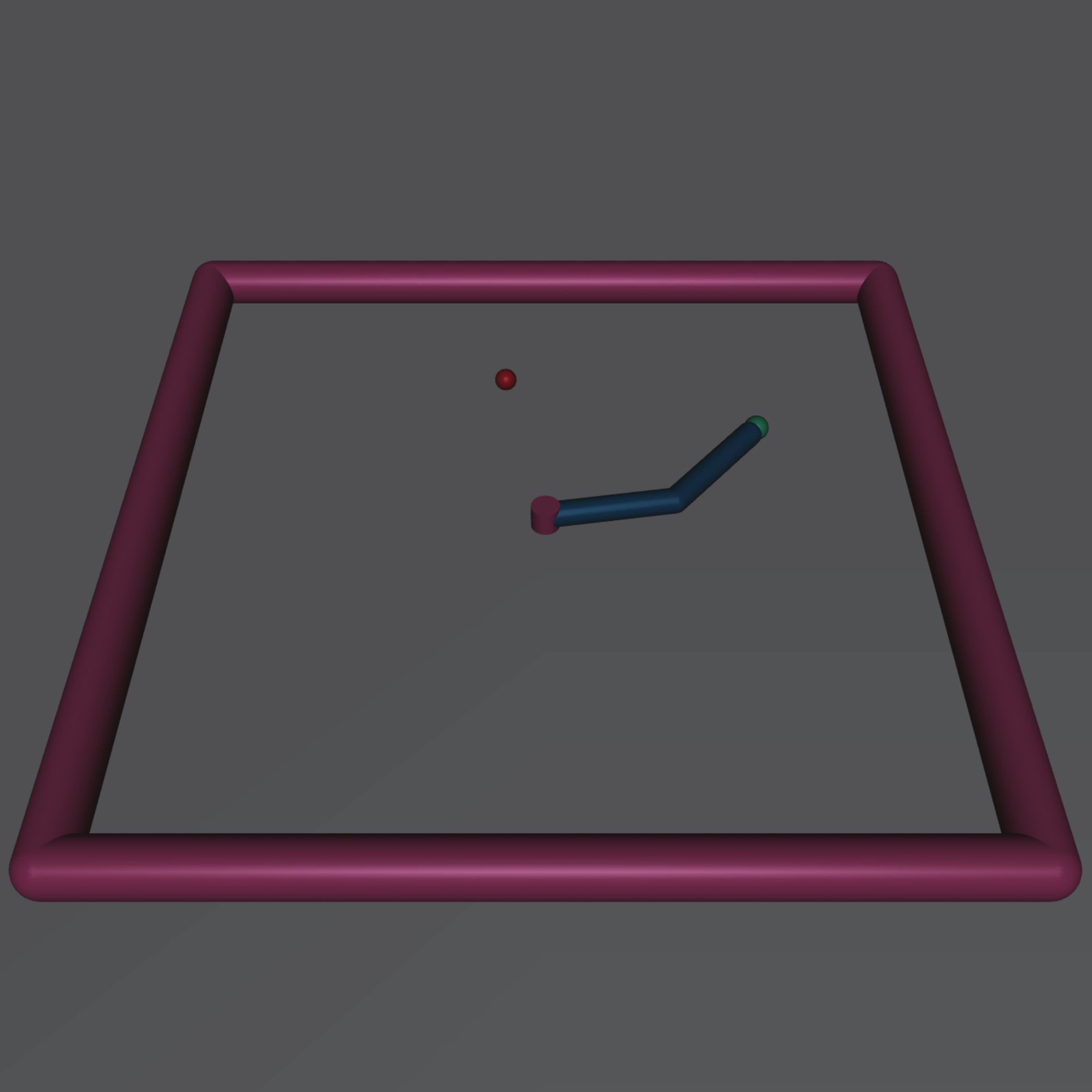} \hfill
  \includegraphics[width=.2428\linewidth]{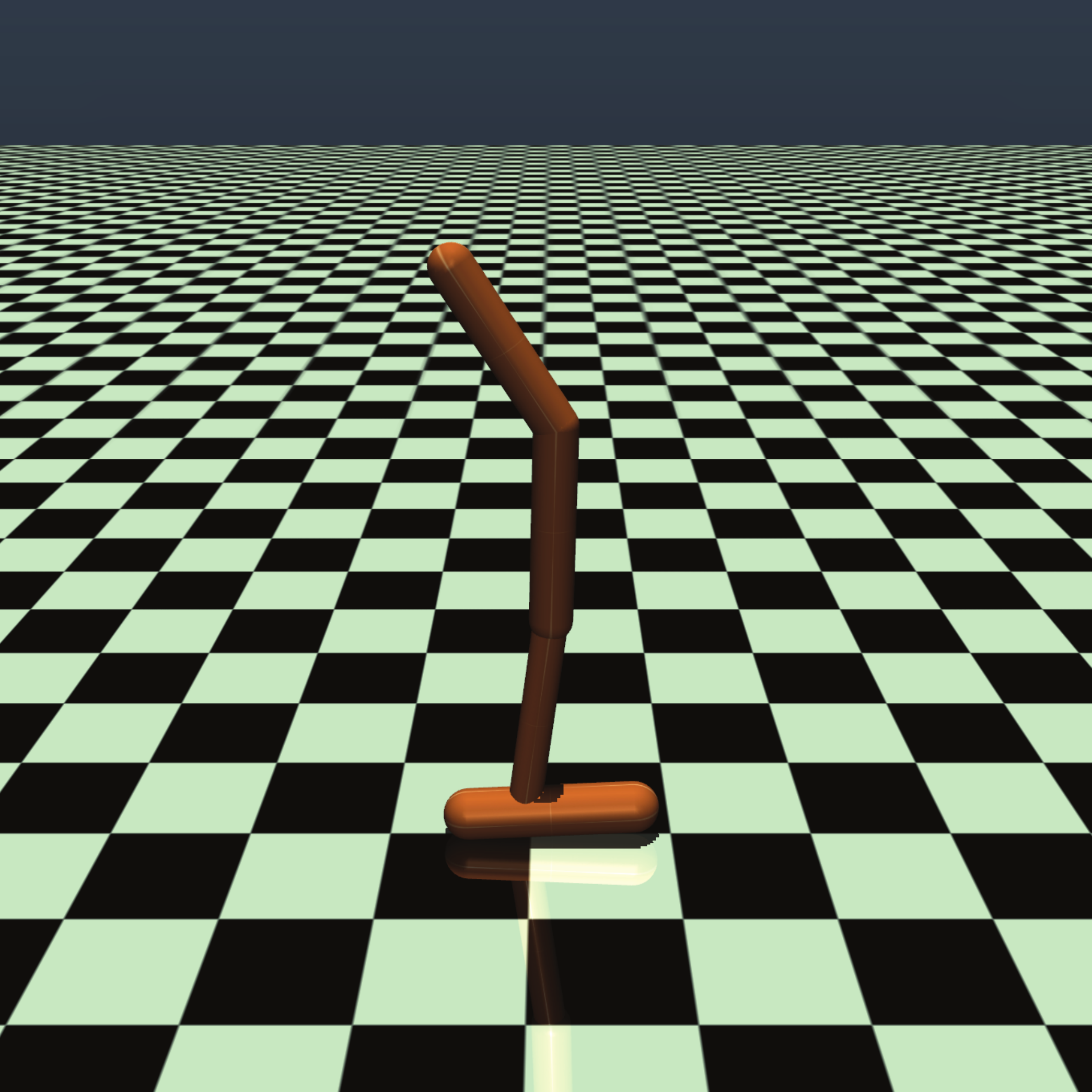} \hfill
  \includegraphics[width=.2428\linewidth]{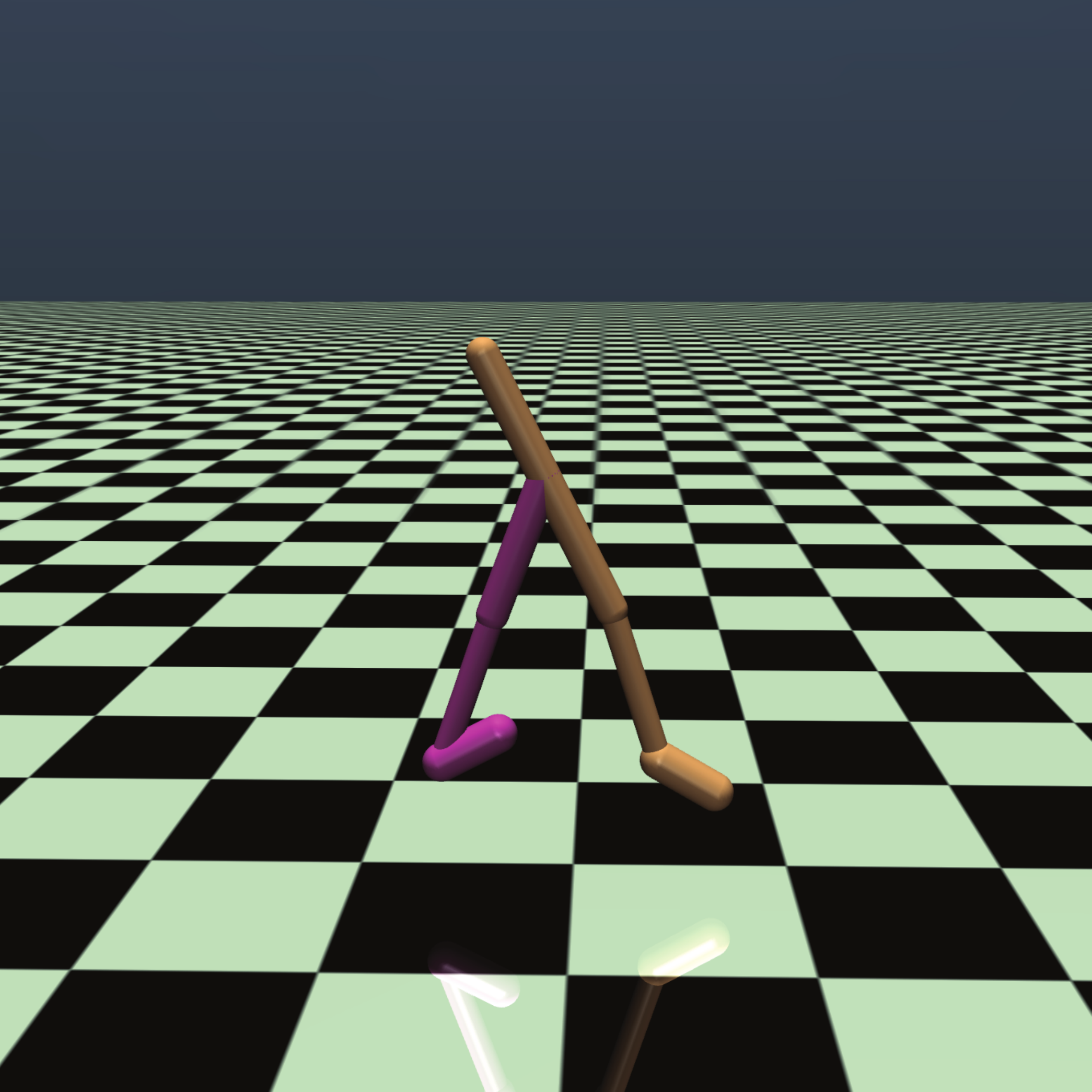} \hfill
  \includegraphics[width=.2428\linewidth]{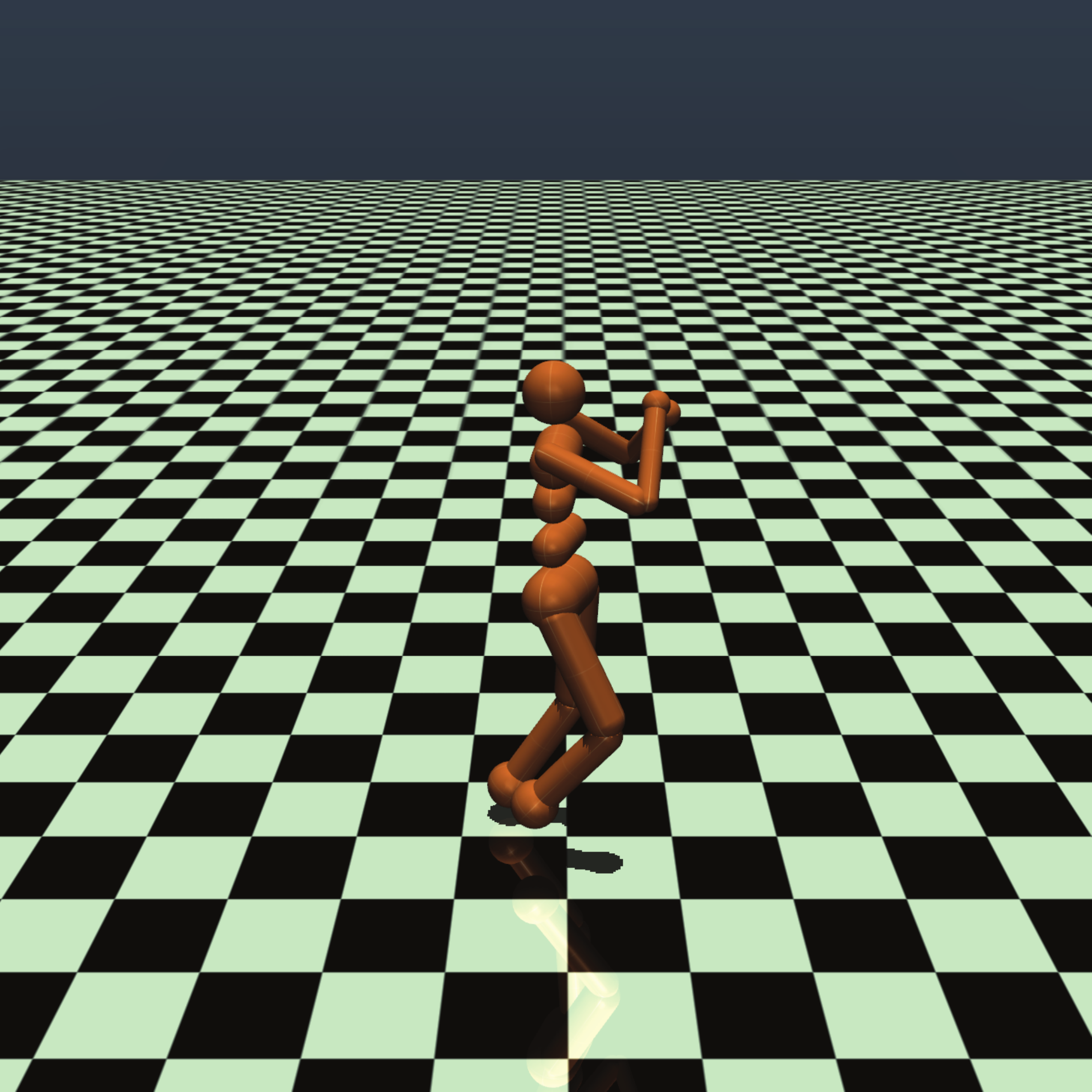}
  \caption{Illustrations of the domains from the OpenAI's MuJoCo Gym that were used in our experiments. From left: Reacher-v1, Hopper-v1, Walker2d-v1, and Humanoid-v1 featuring $2$, $3$, $6$, and $17$ degrees of freedom, respectively.}
\label{fig:illustrations_OpenAI_benchmarks}
\end{figure}

Here we evaluate the performance of BDQ on a set of standard continuous control benchmark domains from the OpenAI's MuJoCo Gym collection. Figure~\ref{fig:illustrations_OpenAI_benchmarks} demonstrates sample illustrations of the environments used in our experiments. Table~\ref{table:dim_specs_OpenAI_benchmarks} states the dimensionality information of these tasks, provided for the specific case of $n=33$ being the finest granularity we experimented with.

We compare the performance of BDQ against a state-of-the-art continuous-action reinforcement learning algorithm, DDPG, as well as against a completely independent agent, IDQ. For all environments, we evaluate the performance of BDQ with two different discretization resolutions resulting in $n=17$ and $n=33$ sub-actions per degree of freedom. We do this to compare the relative performance of BDQ for the same environments with substantially larger discrete action spaces. Where feasible (i.e. Reacher-v1 and Hopper-v1), we also run the Dueling DDQN agent with $n=17$. 

The results demonstrated in Figure~\ref{fig:results_OpenAI_benchmarks} show that IDQ's performance quickly deteriorates with increasing action dimensionality, while BDQ continues to perform competitively against DDPG. Interestingly, BDQ significantly outperforms DDPG in the most challenging domain, the Humanoid-v1 task which involves $17$ action dimensions, leading to a combinatorial action space of approximately $6.5 \times 10^{25}$ possible actions for $n=33$. Our ablation study on BDQ (with a shared network module) and IDQ (no shared network module) verifies the significance of the shared decision module in coordinating the distributed policies, and thus enabling the BDQ agent to progress in learning and to converge to good policies in a stable manner. Furthermore, remarkably, to perform competitively against a state-of-the-art continuous control algorithm in such high-dimensional domains is a feat previously considered intractable for discrete-action algorithms \citep{Lillicrap:2016ddpg,Schulman:2017ppo}. 

However, in the simpler tasks DDPG performs better or on par with BDQ. We think a potential explanation for this could be the use of a specialized exploration noise process by DDPG which, due to its temporally correlated nature, enables effective exploration in domains with momentum.

By comparing the performance of BDQ for $n=17$ and $n=33$, we see that, despite the significant difference in the total number of possible actions, the proposed agent continues to learn rather efficiently and converges to similar final performance levels. An interesting point to note is the exceptional performance of Dueling DDQN for $n=17$ in Reacher-v1. Yet, increasing the action dimensionality by only one degree of freedom (from $N=2$ in Reacher-v1 to $N=3$ in Hopper-v1) renders the Dueling DDQN agent ineffective.      
Finally, it is noteworthy that BDQ is highly robust against the specifications of the TD target and loss function, while it highly deteriorates with the ablation of the prioritized replay. Characterizing the role of the prioritized experience replay, in stabilizing the learning process for action branching networks, remains the subject of future research.      

\begin{table}[t!]
  \begin{center}
  \begin{tabular}{llllllllll}
    \toprule
    Domain  & dim($\boldsymbol{o}$) & $N$ & $n^N$ & $n \times N$ \\
    \midrule
    Reacher-v1 & $11$ & $2$ & $1.1 \times 10^{3}$ & $66$ \\
    Hopper-v1  & $11$ & $3$ & $3.6 \times 10^{4}$ & $99$ \\
    Walker2d-v1 & $17$ & $6$ & $1.3 \times 10^{9}$ & $198$ \\
    Humanoid-v1 & $376$ & $17$ & $6.5 \times 10^{25}$ & $561$ \\
    \bottomrule
  \end{tabular}
  \end{center}
  \caption{Dimensionality of the OpenAI's MuJoCo Gym benchmark domains: dim($\boldsymbol{o}$) denotes the observation dimensions, $N$ is the number of action dimensions, and $n^N$ indicates the number of possible actions in the combinatorial action space, with $n$ denoting the fixed number of discrete sub-actions per action dimension. The rightmost column indicates the total number of network outputs required for the proposed action branching architecture. The values provided are for the most fine-grained discretization case of $n=33$.}
  \label{table:dim_specs_OpenAI_benchmarks}
\end{table}

\section{Experiment Details}
\label{sec:exp_details}

Here we provide information about the technical details and hyperparameters used for training the agents in our experiments. Common to all agents, training always started after the first $10^3$ steps and, thereafter, we ran one step of training at every time step. We did not perform tuning of the reward scaling parameter for either of the algorithms and, instead, used each domain's raw rewards. We used the OpenAI Baselines \citep{baselines} implementation of DQN as the basis for the development of all the DQN-based agents.   

\paragraph{BDQ} We used the Adam optimizer \citep{Kingma:2015adam} with a learning rate of $10^{-4}$, $\beta_1 = 0.9$, and $\beta_2 = 0.999$. We trained with a minibatch size of $64$ and a discount factor  $\gamma =0.99$. The target network was updated every $10^3$ time steps. We used the rectified non-linearity (or ReLU) \citep{Glorot:2011ReLU} for all hidden layers and linear activation on the output layers. The network had two hidden layers with $512$ and $256$ units in the shared network module and one hidden layer per branch with $128$ units. The weights were initialized using the Xavier initialization \citep{Glorot:2010xavier} and the biases were initialized to zero. A gradient clipping of size $10$ was applied. We used the prioritized replay with a buffer size of $10^{6}$, $\alpha = 0.6$, and linear annealing of $\beta$ from $\beta_0 = 0.4$ to $1$ over $2 \times 10^{6}$ steps.   

While an $\epsilon$-greedy policy is often used with Q-learning, random exploration (with an exploration probability) in physical, continuous-action domains can be inefficient. To explore well in physical environments with momentum, such as those in our experiments, DDPG uses an Ornstein-Uhlenbeck process \citep{Uhlenbeck:1930theory} which creates a temporally correlated exploration noise centered around the output of its deterministic policy. The application of such a noise process to discrete-action algorithms is, nevertheless, somewhat non-trivial. For BDQ, we decided to sample actions from a Gaussian distribution with its mean at the greedy actions and with a small fixed standard deviation throughout the training to encourage life-long exploration. We used a fixed standard deviation of $0.2$ during training and zero during evaluation. This exploration strategy yielded a mildly better performance as compared to using an $\epsilon$-greedy policy with a fixed or linearly annealed exploration probability. For the custom reaching domains, however, we used an $\epsilon$-greedy policy with a linearly annealed exploration probability, similar to that commonly used for Dueling DDQN. 

\paragraph{Dueling DDQN} We generally used the same hyperparameters as for BDQ. The gradients from the dueling streams were rescaled by $1/\sqrt{2}$ prior to entering the shared feature module as recommended by \citet{Wang:2016}. Same as the reported best performing agent from \citep{Wang:2016}, the average aggregation method was used to combine the state value and advantages. We experimented with both a Gaussian and an $\epsilon$-greedy exploration policy with a linearly annealed exploration probability, and observed a moderately better performance for the linearly annealed $\epsilon$-greedy strategy. Therefore, in our experiments we used the latter. 

\paragraph{IDQ} Once more, we generally used the same hyperparameters as for BDQ. Similarly, the same number of hidden layers and hidden units per layer were used for each independent network, with the difference being that the first two hidden layers were not shared among the several networks (which was the case for BDQ). The dueling architecture was applied to each network independently (i.e. each network had its own state-value estimator). This agent serves as a baseline for investigating the significance of the shared decision module in the proposed action branching architecture.

\paragraph{DDPG} We used the DDPG implementation of the rllab suite \citep{Duan:2016benchmarking} and the hyperparameters reported by \citet{Lillicrap:2016ddpg}, with the exception of not including a $L_2$ weight decay for Q as opposed to the originally proposed penalty of $10^{-2}$ which deteriorated the performance.

\section{Conclusion}
\label{sec:conclusion}

We introduced a novel neural network architecture that distributes the representation of the policy or the value function over several network branches, meanwhile, maintaining a shared network module for enabling a form of implicit centralized coordination. We adapted the DQN algorithm, along with several of its most notable extensions, into the proposed action branching architecture. We illustrated the effectiveness of the proposed architecture in enabling the application of a currently restricted discrete-action algorithm to domains with high-dimensional discrete or continuous action spaces. This is a feat which was previously thought intractable. We believe that the highly promising performance of the action branching architecture in scaling DQN and its potential generality evoke further theoretical and empirical investigations.

\bibliography{references}
\bibliographystyle{aaai}

\end{document}